\documentclass[10pt,twocolumn,letterpaper]{article}

\usepackage{cvpr}
\usepackage{times}
\usepackage{epsfig}
\usepackage{graphicx}
\usepackage{amsmath}
\usepackage{amssymb}
\usepackage{algorithm}
\usepackage{algorithmic}

\usepackage {subfigure}

\usepackage[pagebackref=true,breaklinks=true,letterpaper=true,colorlinks,bookmarks=true, citecolor=blue]{hyperref}

\cvprfinalcopy 

\def\cvprPaperID{159} 

\ifcvprfinal\fi

\begin{document}

\title{Self-learning Scene-specific Pedestrian Detectors \\
using a Progressive Latent Model}

\author{Qixiang Ye$^{1,4}$, Tianliang Zhang $^{1}$, Qiang Qiu$^{4}$, Baochang Zhang$^{2}$, Jie Chen$^{3}$, and Guillermo Sapiro$^4$
\\
$^1$EECE, University of Chinese Academy of Sciences.\\
$^2$ASEE, Beihang University. $^3$CMV, Oulu University. $^4$ECE,  Duke University.\\
{qxye@ucas.ac.cn}; qixiang.ye@duke.edu }

\maketitle

\begin{abstract}
In this paper, a self-learning approach is proposed towards solving scene-specific pedestrian detection problem without any human' annotation involved. The self-learning approach is deployed as progressive steps of object discovery, object enforcement, and label propagation. In the learning procedure, object locations in each frame are treated as latent variables that are solved with a progressive latent model (PLM). Compared with conventional latent models, the proposed PLM incorporates a spatial regularization term to reduce ambiguities in object proposals and to enforce object localization, and also a graph-based label propagation to discover harder instances in adjacent frames. With the difference of convex (DC) objective functions, PLM can be efficiently optimized with a concave-convex programming and thus guaranteeing the stability of self-learning. Extensive experiments demonstrate that even without annotation the proposed self-learning approach outperforms weakly supervised learning approaches, while achieving comparable performance with transfer learning and fully supervised approaches.

\end{abstract}

\section{Introduction}
With widespread use of surveillance cameras, the need for automatically detecting objects, e.g., pedestrians, has significantly increased. Recent methods ~\cite{Dollar14,Felzenszwalb10,Girshick14,Ren16}
have achieved encouraging progress for detecting objects in images. However, their performance in video scenes is limited for the following main reasons: 1) Supervised learning of detectors for different scenes
requires repeated human effort; 2) Offline-trained detectors usually degrade with changes in the scene or camera; 3) Scene specific cues including object resolution, occlusions, and background structures are not
incorporated into the detectors ~\cite{Shu13}.

\begin{figure}[!t]
\centering
\includegraphics[width=0.50\textwidth]{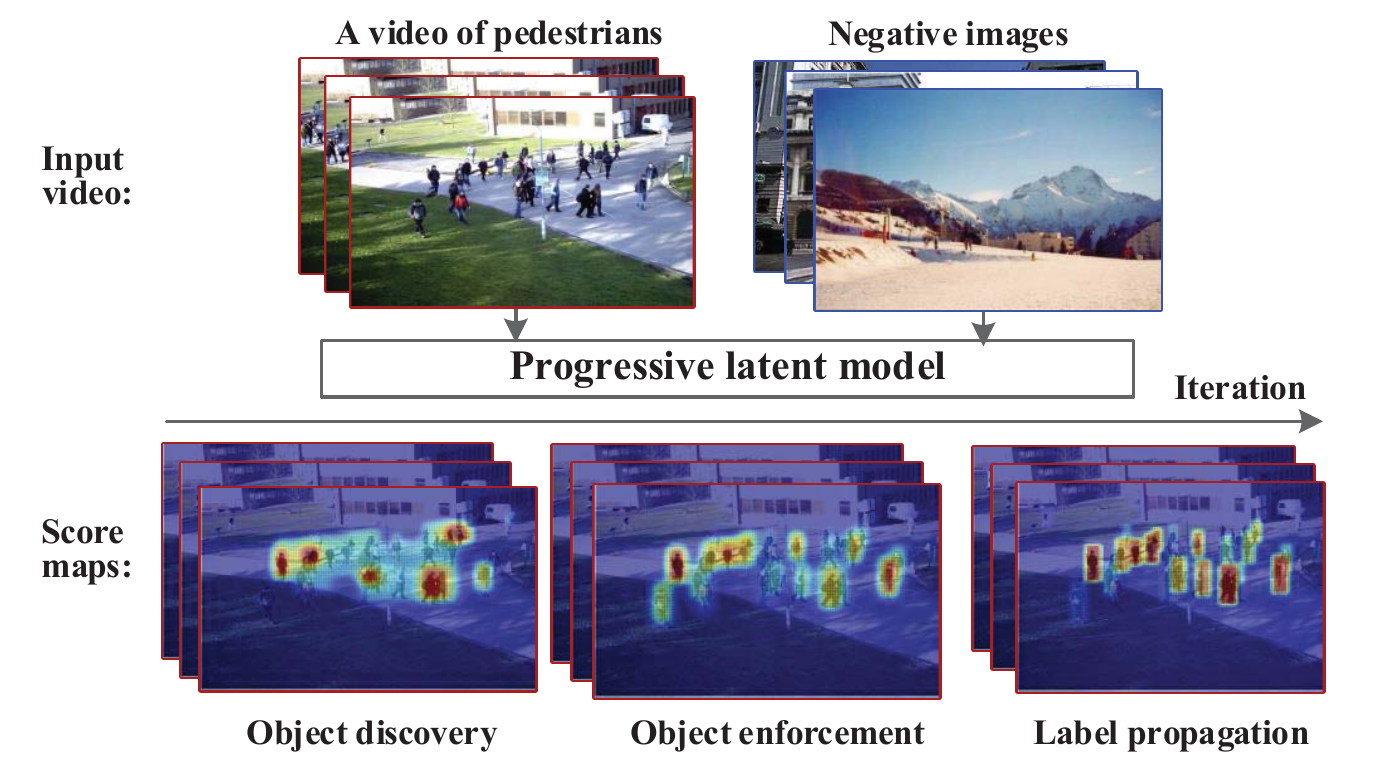}
\caption{\label{Fig.1} \textbf{Proposed self-learning framework}. Given a video where pedestrians are dominant moving objects, self-learning progressively constructs a scene-specific detector using object discovery, object enforcement, and label propagation procedures. }
\end{figure}

Learning scene-specific detectors, which aims at modeling objects in video scenes by incorporating scene-specific discriminative information, has been increasingly investigated
~\cite{Hatori15,Misra15,Stalder09}.
To learn scene-specific detectors with less human supervision, transfer learning and semi-supervised learning are commonly used ~\cite{Hatori15,Misra15,Stalder09}. Transfer learning adapts pre-trained detectors to new specific domains, reduces annotation requirements and improves detector performance~\cite{Wang11,Wang12,Wang14}. Semi-supervised learning saves human annotation effort by initially training detectors with a few annotated examples, and incrementally improving the detectors by extending the sample domains~\cite{Kalal13,Misra15,Yang13}.
However, transfer learning is challenged when the object appearance in the target domains has significant differences with that in the source domains; while semi-supervised models might drift away from the intended aims given noisy or unrelated samples ~\cite{Misra15}.
Most importantly, both methods require partial object-level annotations, and therefore, do not fully eliminate human supervision.

As a promising direction, recent unsupervised video
object discovery techniques ~\cite{Kwak15, Papazoglou13, Xiao16} had been significantly improved, which are supposed to break the bottleneck of the self-taught learning in practical applications. This paper discusses the possibility of self-learning pedestrian detectors in specific and dynamically changing scenes, e.g., a city square, to build a pedestrian detection system in a fully unsupervised manner, given video sequences where pedestrians are the dominant moving objects and additional negative images randomly collected from the Web, Fig.\ \ref{Fig.1}. The problem of self-learning is decomposed into three main components: object discovery, object enforcement, and label propagation. Object discovery is implemented with a latent SVM method \cite{Yu09}, which outputs coarse models and annotations by minimizing frame-level classification error. Object enhancement targets at enforcing object localization and reducing ambiguity, i.e., discriminate object parts with the objects themselves, by leveraging spatial regularization objective. Label propagation optimizes a graph-based objective function to gradually discover harder-positive instances in frames. It also enables the self-learning framework to find complex sample domains, e.g., a manifold space comprising multi-posture and multi-view objects ~\cite{Ye13}. The three procedures are formulated in a progressive latent model (PLM) with difference of convex (DC) objective functions, which are efficiently optimized with concave-convex programming in a progressive manner.

The main contributions of this paper consist of:
(1) A self-learning pedestrian detection framework, which is deployed as iterative procedures of object discovery, object enforcement and label propagation, posing a new direction in the field of (unsupervised) object detection;
(2) A progressive latent model (PLM), which uses spatial-temporal regularization to reduce ambiguity of discovered samples, as well as addressing the stability of self-learning; and
(3) Extensive experiments on PETS2009, Towncenter, PNN-Parking-Lot2/Pizza, CUHK Square, and 24-Hours datasets are conducted to verify the performance of the proposed approach.


\section{Related Works}

Pedestrian detection using supervised methods has been extensively investigated~\cite{Cai2015,Dollarr12,Kewei15,Tian2015,Ye13,Zhang16}. This work, however, is more related to scene-specific detection using transfer learning, online learning, weakly supervised learning, and unsupervised object discovery.

\textbf{Transfer learning:} The motivation behind transfer learning is that contexts and object distributions in target domains might be leveraged to improve the performance of pre-trained
detectors in source domains. Researchers have explored context cues ~\cite{Wang11,Wang14}, confidence propagation ~\cite{Wang14,Zeng14}, and virtual-real world adaptation ~\cite{Vazquez14} to realize smooth transfer. Gaussian process regression ~\cite{Xu14} and super-pixel region clustering ~\cite{Shu13} have been explored to select ``safe" samples in target domains. Large margin embedding
~\cite{Kuznetsova15} and transductive multi-view embedding ~\cite{Fu14} have been explored to expand detector horizons. Researchers have also been using domain adaptation to construct a self-learning-camera
~\cite{Gaidon14}.

Transfer learning can obviously reduce human annotations. Nevertheless, it suffers from the concept gap problem, i.e., the major differences of object appearance, viewpoint, and illumination between source and target domains. When the gap is significant, the adaptation of pre-trained models becomes non-smooth or infeasible. By contrast, self-learning initializes and improves detectors in the same scenes, naturally avoiding the concept gap problem.

\textbf{Online/semi-supervised learning}:
Online learning and semi-supervised learning improves scene-specific detectors by taking advantage of the continuous incoming data stream from the target domains. 
Classical detection-by-tracking (DBT) ~\cite{Andriluka08,Mao15} initializes the system using offline trained detectors and leverages temporal cues to extend sample domains and cancel detection errors. Tracking-Learning-Detection (TLD) ~\cite{Kalal12} initializes the system with a single sample, and uses tracking and online learning to boost detectors. Despite the popularity of DBT and TLD approaches, recent studies ~\cite{Misra15} demonstrated that the simple combination of detection with tracking might introduce poor detectors because the errors from both detection and tracking could be amplified in a coupled system.
A P-N expert ~\cite{Kalal12} is used in TLD to control precision and recall rates, guaranteeing the learning stability as a linear dynamic system. The learning stability of our approach can also be guaranteed as the difference of convex (DC) objective functions of PLM converge at each learning iteration.

\textbf{Weakly supervised learning:} The inputs of WSL are image/video level tags (object category), and the algorithm discovers objects when learning detectors ~\cite{Kwak15, Song14}. A
general assumption behind WSL is that objects of the same category are from a potential cluster while the backgrounds are diverse. Under such an assumption, clustering ~\cite{Divvala15,Chong14}, tracking
~\cite{Kwak15}, boosting ~\cite{Wu07}, region matching ~\cite{Cho115}, graph labeling ~\cite{Song14}, and multi-instance learning ~\cite{Cinbis16,Ren2016} are used to find the correspondence of objects, depress
the backgrounds and learn detectors.

WSL alternates between sample labeling and detector learning
in a way similar to Expectation Maximization optimization. Due to the missing annotations, however, this optimization is non-convex and therefore prone to getting stuck in a local minimum and outputting wrong
labelings ~\cite{Bilen15}.
Cinbis \textit{et al.} ~\cite{Cinbis16} use a multi-fold splitting of the training set while Bilen \textit{et al.} ~\cite{Bilen15} use convex clustering to prevent getting stuck to wrong labels.
This work alleviates the local optima problem with a more reasonable way by introducing regularization terms about domain knowledge, i.e., intra-frame hard-negative mining and inter-frame similarity propagation.

\textbf{Unsupervised video object discovery:}
An early approach developed in ~\cite{Wu07} learns scene-specific object detector by online boosting of part detectors, but it requires general seed detectors learned offline. Recent research ~\cite{Kwak15,Xiao16}
formulates unsupervised video object discovery as a combination of two complementary steps: discovery and tracking. The first step establishes correspondences between prominent regions across video frames, and the second step associates successive similar object regions within the same video. Xiao \textit{et al.} ~\cite{Xiao16} propose a fully unsupervised video object proposal approach which first discovers a set of easy-to-group instances by clustering and then updates its appearance model to gradually detect harder instances by the initial detector and temporal consistency. This unsupervised approach can automatically generate object proposals, but cannot output precise detections.

\section{Proposed Self-learning Framework}

In the supervised object detection setting, the locations of training samples would simply be given, while in self-learning, the annotations of object locations are not available. The primary objective of self-learning is guiding the missing annotations to a solution
that disentangles object samples from noisy object proposals, as shown in Fig.~\ref{Fig.2}.

\subsection{Progressive Latent Model}

\begin{figure}[!t]
\centering
\includegraphics[width=0.42\textwidth]{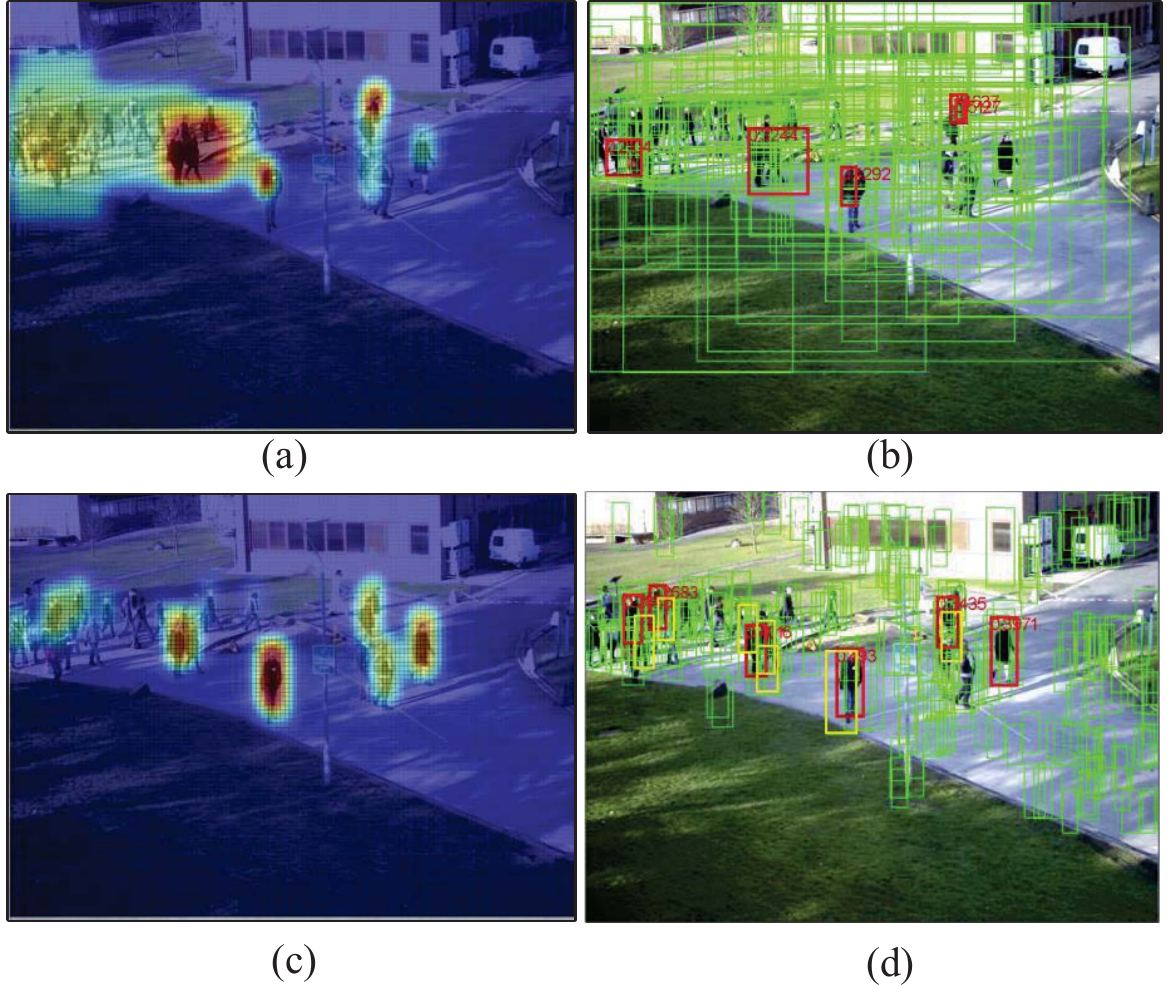}
\caption{\label{Fig.2} Object discovery from noisy proposals. (a) The score map in the first learning iteration and (b) candidate objects (red boxes) discovered. (c) The score map and in the fifth learning iteration. (d) Candidate objects (red boxes) and hard negatives (yellow boxes). (Best viewed in color.)}
\end{figure}

\textbf{Modeling:}
The self-learning framework is decomposed into three basic procedures: object discovery, object enhancement, and label propagation. Given a set of object proposals that have salient object-like appearance and motion,  Fig.~\ref{Fig.2}a and Fig.~\ref{Fig.2}b, the object discovery step aims to find object windows from video frames that best discriminates positive video frames from the negative images. The object enhancement discovers hard negatives that help reducing falsely localized object parts, as well as improving object localization. The label propagation step mines harder instances of the corresponding object and throughout the entire video, Fig.~\ref{Fig.2}c and Fig.~\ref{Fig.2}d. The three procedures iterate until an error rate based stability criteria is met.

Let $x\in \mathcal{X}$ denotes a video frame or a negative image, $y\in \mathcal{Y}, \mathcal{Y}=\{0,1\}$ are labels denoting if $x$ contains a pedestrian object. $y=1$ indicates that there is at least one pedestrian in the frame while $y=0$ indicates a frame without pedestrian object or a negative image. The self-learning is formulated with a multi-objective function that targets at jointly determining the latent object $h$ and a latent model $\beta$ in a progressive optimized procedure,

\begin{equation}\label{Eq.1}
\begin{aligned} \left\{{h^*},{\beta^*} \right\} &= \min_{\beta, h}\mathcal{F}_{(\mathcal{X,Y})}(\beta,h) \\
                                                &=\min_{\beta,h} \mathcal{F}_l(\beta,h) - \lambda \mathcal{F}_s(\beta) +\gamma \mathcal{F}_g(\beta,h),
\end{aligned}
\end{equation}
where $\mathcal{F}_l(\beta, h)$, $\mathcal{F}_s(\beta)$ and $\mathcal{F}_g(\beta, h)$ \footnote{$(\mathcal{X,Y})$ is omitted for short.}, as defined below, are the objectives for object discovery, spatial regularization and score propagation respectively. $\lambda$ and $\gamma$ are regularization factors.

\textbf{Object Discovery:}
The object discovery procedure is implemented with a latent SVM (LSVM) model to choose object proposals that best discriminate positive frames from negative images,
\begin{equation}\label{Eq.2}
\left\{ {{y^*},{h^*},{\beta^*}} \right\} = \mathop {\arg \max }\limits_{y \in \mathcal{Y}, h \in \mathcal{H}, \beta} \;\beta^T \cdot v \left( {x,y,h} \right),
\end{equation}
where $v(x,y,h)$ denotes a normalized feature vector, i.e., HOG features. $\mathcal{H}$ denotes the set of object proposals, made up of proposals $\mathcal{H}_i, i=1,...,N$ from video frames.
Basically, solving Eq.~\ref{Eq.2} produces a high score $\beta^T \cdot v(x,y,h)$ for each positive frame $(y=1)$ and a low score for each negative image $(y=0)$. Concretely, we learn the model $\beta$ on a collection of video frames and negative images $X = \{(x_i, y_i), i=1,...,N\}$ with
\begin{equation}\label{Eq.3}
\begin{aligned}
\min_{\beta, h}\mathcal{F}_l(\beta,h) = \min_{\beta,h}\frac{1}{2}||\beta||^2+\mathcal{C}\sum_{i=1}^{N}l(\beta,x_i,y_i,h),
\end {aligned}
\end{equation}
where $\mathcal{C}$ is a regularization factor and $l$ is a difference-convex loss function defined as
\begin{equation}\label{Eq.4}
\begin{aligned}
l(\beta,x_i,y_i,h) = &\max_{y,h}\big(\beta ^T\cdot v(x_i,y,h)+\Delta(y_i,y)\big) \\
&- \max_{h}\beta^T \cdot v(x_i,y_i,h),
\end {aligned}
\end{equation}
where $\Delta(y_i,y)=0$ if $y=y_i$, and 1 otherwise. Eqs. \ref{Eq.3} and \ref{Eq.4} target at choosing and discriminating the highest scoring proposals $h$ from the other configurations, defining a max-margin formulation to measure the mismatch between the image, label, and proposals.

\textbf{Object Enforcement:} The object discovery procedure aims at optimizing the image-level classification instead of the sample-level classification. Once the image-level classification objective function reaches optimization, whether or not the sample-level classification is optimized, the learning procedure stops \cite{Yu09}. Considering that all positive images contain the object parts but none of negative images does, LSVM could falsely select object parts as ¡°positive¡± samples since Eq.~\ref{Eq.3} is non-convex and is easy to get stuck to local minimum.

Motivated by the success of hard negative mining ~\cite{Girshick15}, we propose using spatial regularization to enforce the localization of objects and the model.
Denoting by $\mathcal{H}_i$  object proposals in frame $i$ and $h'$ the hard negatives corresponding to an object $h$ in a video frame, we define a function to maximize the distance between the potential object and its spatial neighbors,
\begin{equation}\label{Eq.5}
\begin{aligned}
\max_{\beta}\mathcal{F}_s(\beta) \! = \!\sum_{i=1}^{N} \! \sum_{h\in\mathcal{H}_i \atop{h'\in\Omega_{\mathcal{H}_i,h}}} \! ||\beta^T \!\cdot \!\big( v(x_i,h)\!-\!v(x_i,h')\big) ||^2,
\end {aligned}
\end{equation}
where $\Omega_{\mathcal{H}_i,h}$ denote the spatial neighbors of $h$ in $\mathcal{H}_i$.
The spatial neighbors are high score object parts and surrounding image patches that have IoU (Intersection of Union) with $h$ in the interval (0.0 0.25). Eq.~\ref{Eq.5} optimizes the model $\beta$ using fixed $h$, and thus is a convex regularization function. Such a function enforces the latent model, yielding a consistent and significant boosts in object localization with a progressive learning procedure.

\textbf{Label Propagation:} 
The object discovery procedure outputs only one sample for each frame. To mine more positives and negatives, we propose using the inter-frame label propagation for incremental learning.

Suppose there are $l$ labeled samples from previous learning iterations. We select $u=l\times(r-1.0)$ high-scored proposals as unlabeled samples, where $r>1.0$ is the learning rate, related to the expected density of pedestrians.
Given labeled samples $\{h_i\}, i=1,...,l$, and unlabeled proposals $\{h_j\}, j=l,...,l+u$, a $k$NN graph in the feature space is first constructed. The graph vertex defines the nearest neighbor vertices of samples. $h_i$ and $h_j$ are connected if one of them is among the other¡¯s $k$NN ~\cite{Zhu09}. The graph-based label propagation
procedure is defined as  $g(\beta,h_j)=\frac{\sum_{k=l}^{l}w_{jk}g(\beta,h_k)}{\sum_{k=l}^{l}w_{jk}},j=l+1,...,l+u,$
where $w_{ik}$ denotes the edge weight defined with a Gaussian Function on Euclidean distance between $h_i$ and $h_k$. This is equivalent to a convex optimal problem \cite{Zhu09},
\begin{equation}\label{Eq.6}
\begin{aligned}
\min_{g(\beta,h)}\mathcal{F}_g(\beta,h) = &\min_{g(\beta,h)}\sum_{i=1}^{l}\sum_{j=l}^{l+u}w_{ij}\big(g(\beta,h_i)-g(\beta,h_j)\big)^2\\
&s.t.\quad g(\beta,h_i)=y_i, i=1,...,l,
\end{aligned}
\end{equation}
where $g(\beta,h_j)$ is the propagated score of proposal $h_j$ and $y_i$ is the label of the frame/image that $h_i$ belongs to.
%

\textbf{Progressive Optimization:}
In the learning procedure, the optimization of $F_s(\beta)$ (object enforcement) and $F_g(\beta,h)$ (label propagation) depends on the results of $F_l(\beta,h)$. Eq.\ \ref{Eq.1} is thus a progressive model, where $F_l$ , $F_s$ and  $F_g$ are alternatively optimized. According to Eq.\ \ref{Eq.4}, $\mathcal{F}_l$ could be written as $A(x)-B(x)$ and $\mathcal{F}$ could be written as  $A(x)-B(x)+C(x)-D(x)$. This means that the objective functions of Eq.\ \ref{Eq.1} could be written as the difference of convex functions. This allows us to optimize it with a two-step Concave-Convex Procedure (CCCP) \cite{Yu09}. The first-step CCCP for $\mathcal{F}_l$  discovers potential pedestrian objects in frames and initializes the latent model, the second-step CCCP for $\gamma\mathcal{F}_g-\lambda\mathcal{F}_s$ performs object enforcement and label propagation. The two-steps CCCP progressively optimizes the PLM until the change of the estimated sample error rate is negligible. CCCP algorithms guarantee the optimization with difference of convex objective functions converges to a local minimum or saddle point \cite{Yu09}. Therefore, iterative usage of the two-steps CCCP algorithm and keeping the decreasing of the sample error rates (discussed in Sec.\ \ref{sec:ErrorRate}) can guarantee the stability of self-learning.

\subsection{Self-learning a Detector}
\begin{figure}[!t]
\centering
\includegraphics[width=0.48\textwidth]{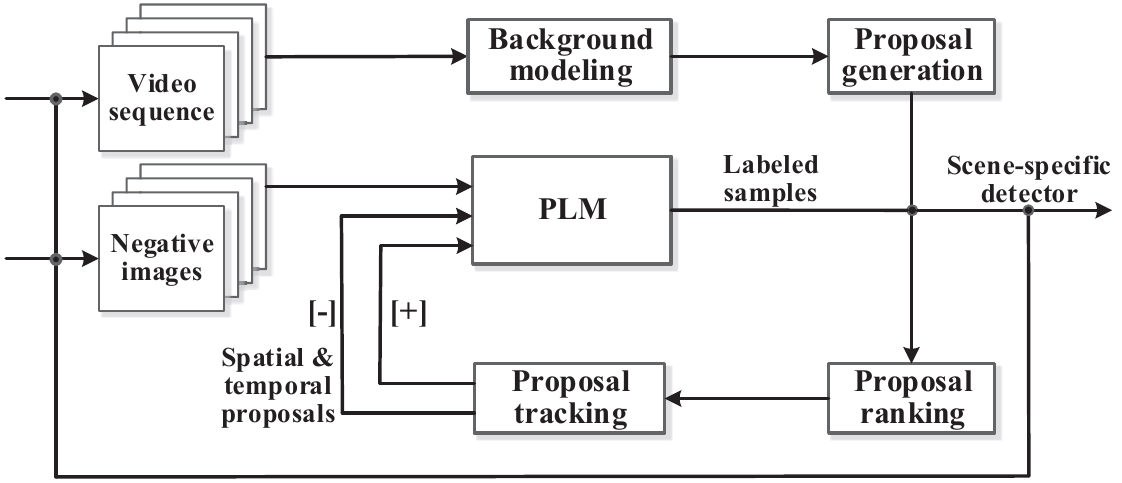}
\caption{\label{Fig.3} Block diagram of the proposed self-learning approach}
\end{figure}

With the proposed PLM, a self-learning approach is implemented as described in Fig.\ \ref{Fig.3}. The proposal generation component localizes potential objects using objectness, motion, and appearance cues. The proposal ranking component chooses the high-ranked
proposals as positive candidates, and low-ranked proposals as negatives. The proposal tracking component helps in finding proposals in successive video frames. The PLM identifies positives and hard negatives from given proposals. With mined positive samples, a DPM detector $f_\beta(h)$ is trained to perform pedestrian detection.


Given a video of static background, a motion score map is calculated for each video frame with a background modeling algorithm. On the motion score map, detection proposals (as shown in Fig.~\ref{Fig.2}b) are extracted using the EdgeBoxes approach ~\cite{Zitnick14}, according to which edge maps are computed first, and contours, i.e., edge groups, are obtained by aggregating high affinity edges. On the contours, the regions of high confidence are extracted as object proposals using a sliding window strategy in locations, scales, and aspect ratios.
From the second iteration,  with an initialized detector, a sliding window strategy is used to generated object proposals, as shown in Fig.~\ref{Fig.2}d. To extend the proposals in the temporal
domain, a KLT tracking algorithm is employed to track and collect proposals from frame $t$ to frame ${t+\tau}$, where $\tau$ is empirically set to 10. Before feeding these spatial-temporal proposals to the learning algorithm, their aspect ratios are normalized to the average aspect ratio.
To prevent falsely choosing static backgrounds in videos of sparse pedestrians, the average background probability of a proposal is required to be larger than a threshold, empirically set to 0.20 in our experiments.

We propose using a combinatorial score, i.e., $f(h) = \alpha^T\cdot(f_\beta(h),f_m(h),f_o(h)),$ to choose high-ranked proposals, where $\alpha^T$ is a ranking weight vector.  $f_\beta(x)$, $f_m(h)$ and $f_o(h)$,
respectively, are the detection, motion, and objectness scores. The motion score $f_m(h)$ of a proposal is defined as the averaged motion scores of all pixels in its image region. Objectness score $f_o(h)$ is defined by calculating contours in the proposal regions ~\cite{Zitnick14}. A larger score gives higher confidence that the proposal is an object. Detection score $f_\beta(h)$ is calculated from the second learning iteration, by the learned detector. From this iteration, the proposal region centers are set as root locations, around which we use sliding window to localize proposals.

In each learning iteration, the ranking weight vector $\alpha^T$ is updated using a zero-space regression method ~\cite{Chang11}, which performs learning without using output
values. It basically minimizes the regression error of all samples, as well as maximizing the distance from a hyperplane to the origin. This results in a weight vector which captures regions in the input sample
space where the probability density of the data is found, and enables the proposal ranking to be adaptive.

\subsection {Error Rate Discussion} \label{sec:ErrorRate}
PLM incorporates a label propagation procedure, which iteratively introduces new samples and updates the model. In this procedure, the primary problems to be solved are avoiding model drift and reducing the error rate. Eq.\ \ref{Eq.6} implies that a larger $\gamma$ value introduces more newly labeled samples, as well as a larger error rate $\xi$, and vice versa. The number of newly labeled samples $u$ is determined to be an implicit function of $\gamma$, $u(\gamma)$.
The value of $\gamma$  needs to essentially guarantee that the error rate of newly labeled samples is smaller than that of existing samples, meaning the error rate of the training set is monotonically non-increased.
It is also expected that there is a large $\gamma$, which implies that more samples could be labeled in each iteration. To decide the value of $\gamma$, an optimization objective function is defined:

\begin{equation}\label{Eq.7}
\begin{aligned}
&\max_{\gamma,\beta,y_j}\gamma \\
&s.t.\quad\xi_{u(\gamma)} \leq \xi_l \\
&\approxeq\frac{1}{l+u(\gamma)}\sum_{j=1}^{l+u(\lambda)}(f_\beta(h_j)-\widetilde y_j)\leq\frac{1}{l}\sum_{i=1}^{l}(f_\beta(h_i)-\widetilde y_i),\\
\end{aligned}
\end{equation}
where $l$ and $u(\gamma)$, respectively, denote the numbers of labeled samples in previous iterations and unlabeled samples in current iteration.

The optimization of Eq. \ref{Eq.7} guarantees that the estimated error rate of newly labeled samples $\xi_{u(\gamma)}$ is smaller than that of labeled samples $\xi_l$ by finding a proper $\gamma$ in each learning iteration. $\gamma$ is optimized with a linear searching algorithm ~\cite{Donald97}, which searches in the interval [0.0, 1.0] with step size 0.1 and updates $f_\beta(h_j)$ to $f_{\widetilde\beta}(h_j)$ at each step. Meanwhile, $\widetilde y_j$ is estimated with $\widetilde y_j = f_{\widetilde\beta}(\cdot)$, with which the error rate $\xi_{u(\gamma)}$ is calculated.


%

\section{Experiments}
\subsection{Datasets and Performance Metrics}
The proposed approach is evaluated on five real-world datasets (six sequences) captured with surveillance cameras. The datasets involve challenges from object occlusions, low resolution, and/or moving distractors\footnote{A demo video has been included in the supplementary materials.}.


\noindent\textbf{PETS2009~\cite{Ferryman09}:} A crowded video sequence captured in a public space, with 720$\times$576 resolution.

\noindent\textbf{Towncenter~\cite{Benfold11}:} A moderately crowded video sequence of a town center, with 1920$\times$1080 resolution.

\noindent\textbf {PNN-Parking-Lot2/Pizza~\cite{Shu13}:} Moderately crowded video sequences including groups of pedestrians walking in queues with complex motion and similar appearance, with 1920$\times$1080 resolution. It is challenging due to the large amounts of pose variations and occlusions.

\noindent\textbf{CUHK Square~\cite{Wang14}:} A 60-minutes long video of sparse pedestrians and other moving distractors, e.g., moving vehicles. The resolution of the video is 704$\times$576. The resolution of pedestrian objects is much lower than those of other datasets. As the camera has an approximately 45-degree bird-view, objects have perspective deformation.

\noindent\textbf{24Hours:~\footnote{It will be a publicly available dataset.}} A 24-hours long video of sparse/dense pedestrians, 24-hour illumination change and other moving distractors, e.g., moving vehicles, which allows to asses model drift. The resolution of the video is 704$\times$576. 6000 frames were uniformly sampled from the long video for learning and 2600 frames for testing.

For all datasets except the 24Hours, half of the video frames are used for learning while the other annotated frames are used for testing. The proposed approach is evaluated and compared against the following supervised learning, transfer learning, and weakly supervised learning approaches.

\noindent\textbf{Offline-DPM~\cite{Felzenszwalb10}:} A DPM detector off-line trained on the PASCAL VOC person class.

\noindent\textbf{Supervised-DPM:} A supervised DPM detector trained with human annotated samples on specific scenes and additional negative samples mined from negative images.

\noindent\textbf{Supervised-SLSV~\cite{Hatori15}:} A state-of-the-art scene-specific pedestrian detector learned from virtual pedestrians whose appearance is simulated in the specific scene under consideration. Without public available source code, SLSV is only compared on the Towncenter dataset using author reported results.

\noindent\textbf{Transfer-DPM~\cite{Shu13}:} A scene-specific detection approach based on transfer learning. Detections are originally obtained with a DPM detector off-line trained using PASCAL VOC person class and then improved using super-pixel based clustering and classification.

\noindent\textbf{Transfer-SSPD~\cite{Wang14}:} A state-of-the-art scene-specific pedestrian detector with transfer learning.

\noindent\textbf{Weakly-MIL~\cite{Cinbis16}:} A widely used weakly supervised approach based on multi-instance learning. A DPM learner is then learned from annotated positive samples.

\begin{figure}
\centering
\subfigure[]{
\begin{minipage}[b]{0.22\textwidth}
\includegraphics[width=1\textwidth,height=0.8\textwidth]{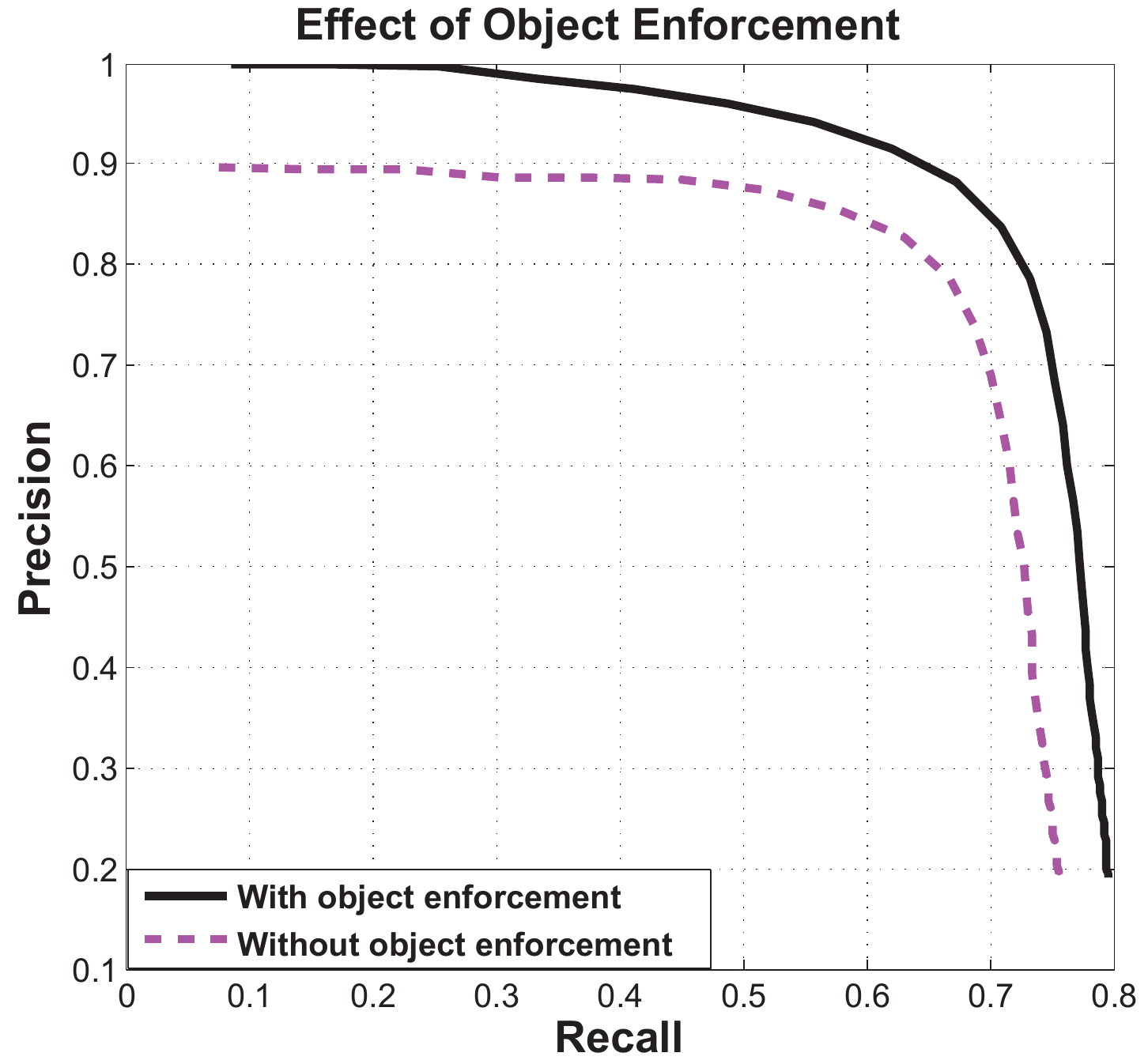}
\end{minipage}
}
\subfigure[]{
\begin{minipage}[b]{0.22\textwidth}
\includegraphics[width=1\textwidth,height=0.8\textwidth]{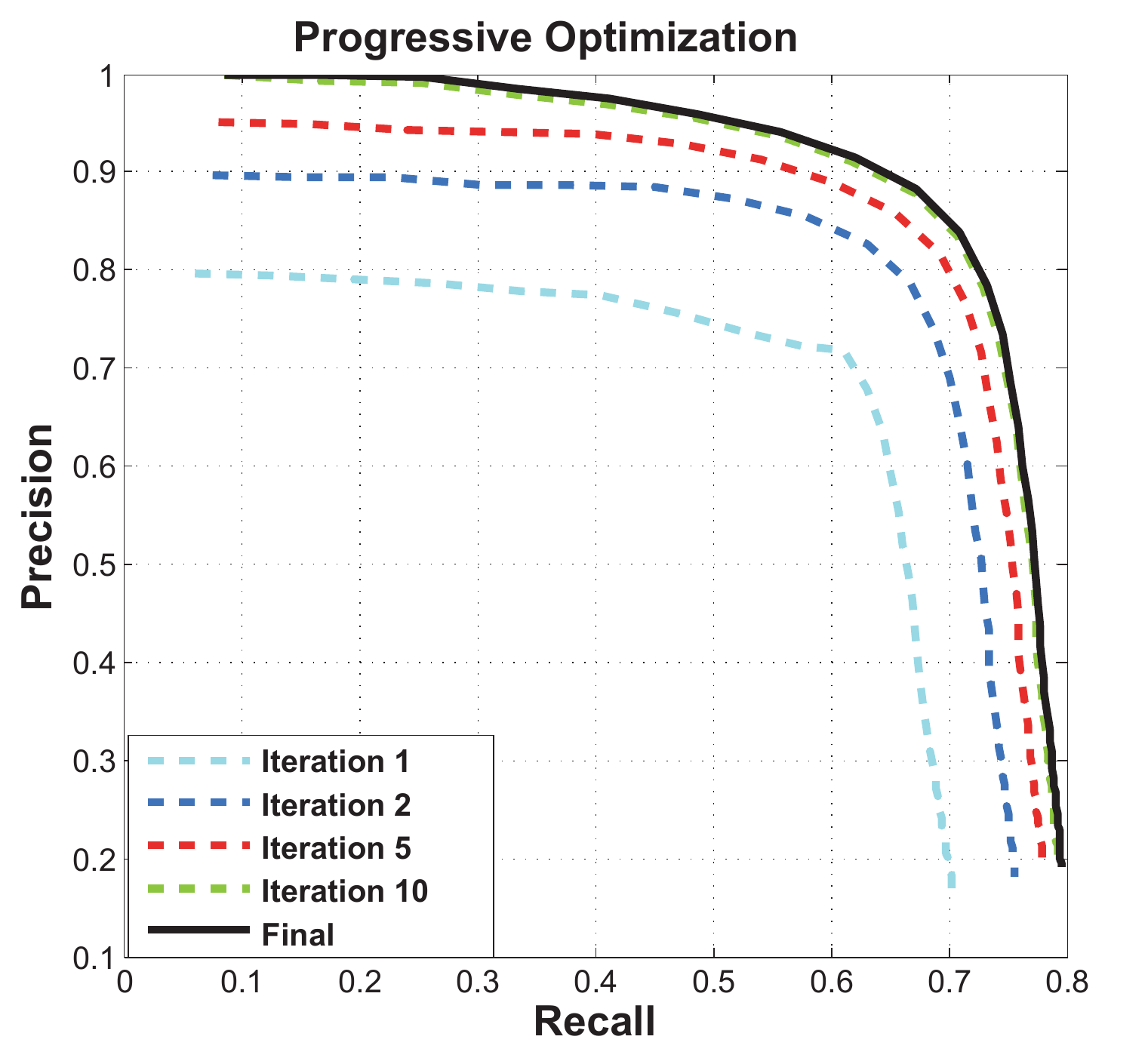}
\end{minipage}
}
\caption{\label{Fig.4} Model effect.
}
\end{figure}

\begin{figure}
\centering
\subfigure[]{
\begin{minipage}[b]{0.22\textwidth}
\includegraphics[width=1\textwidth,height=0.8\textwidth]{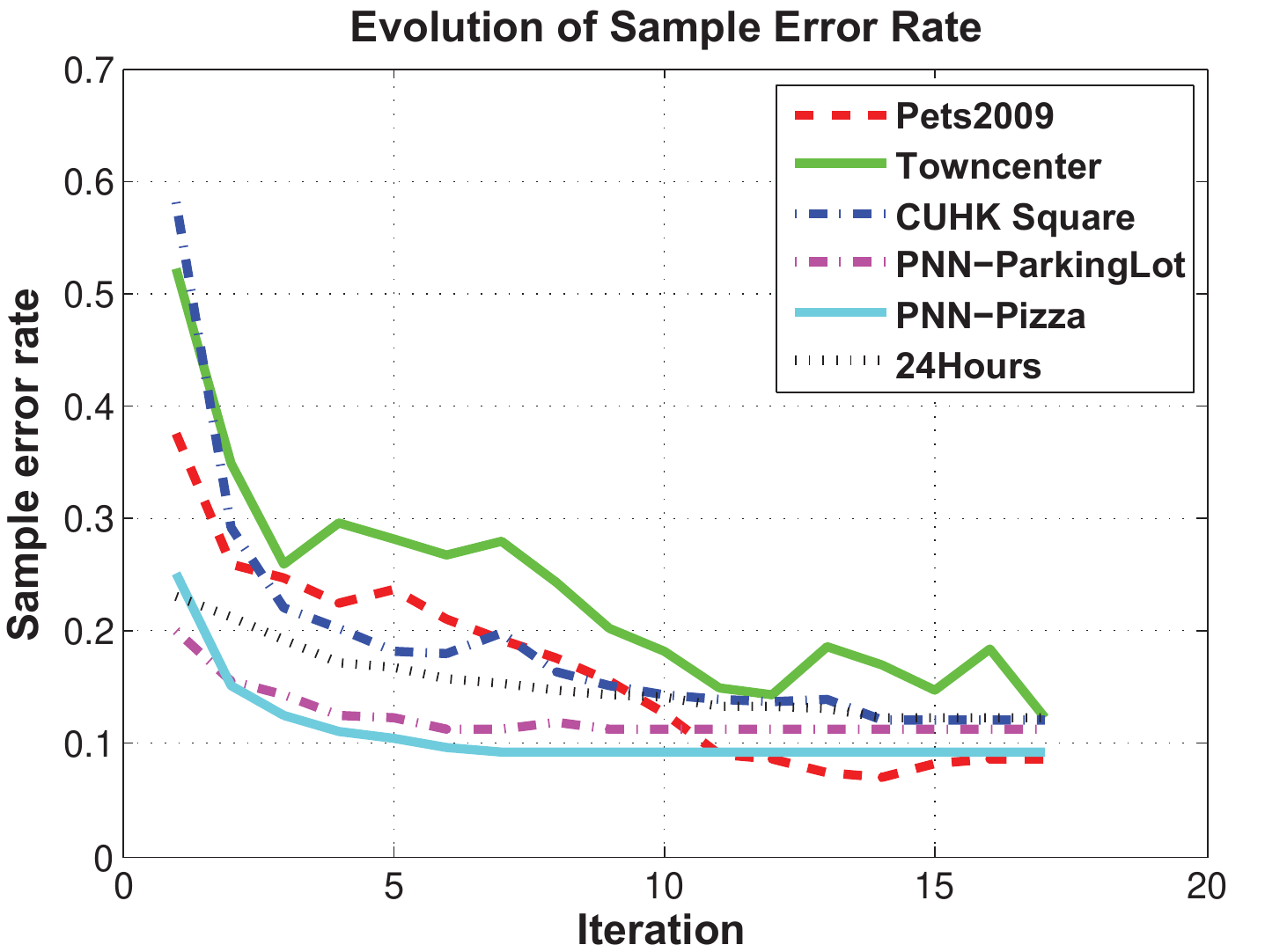}
\end{minipage}
}
\subfigure[]{
\begin{minipage}[b]{0.22\textwidth}
\includegraphics[width=1\textwidth,height=0.8\textwidth]{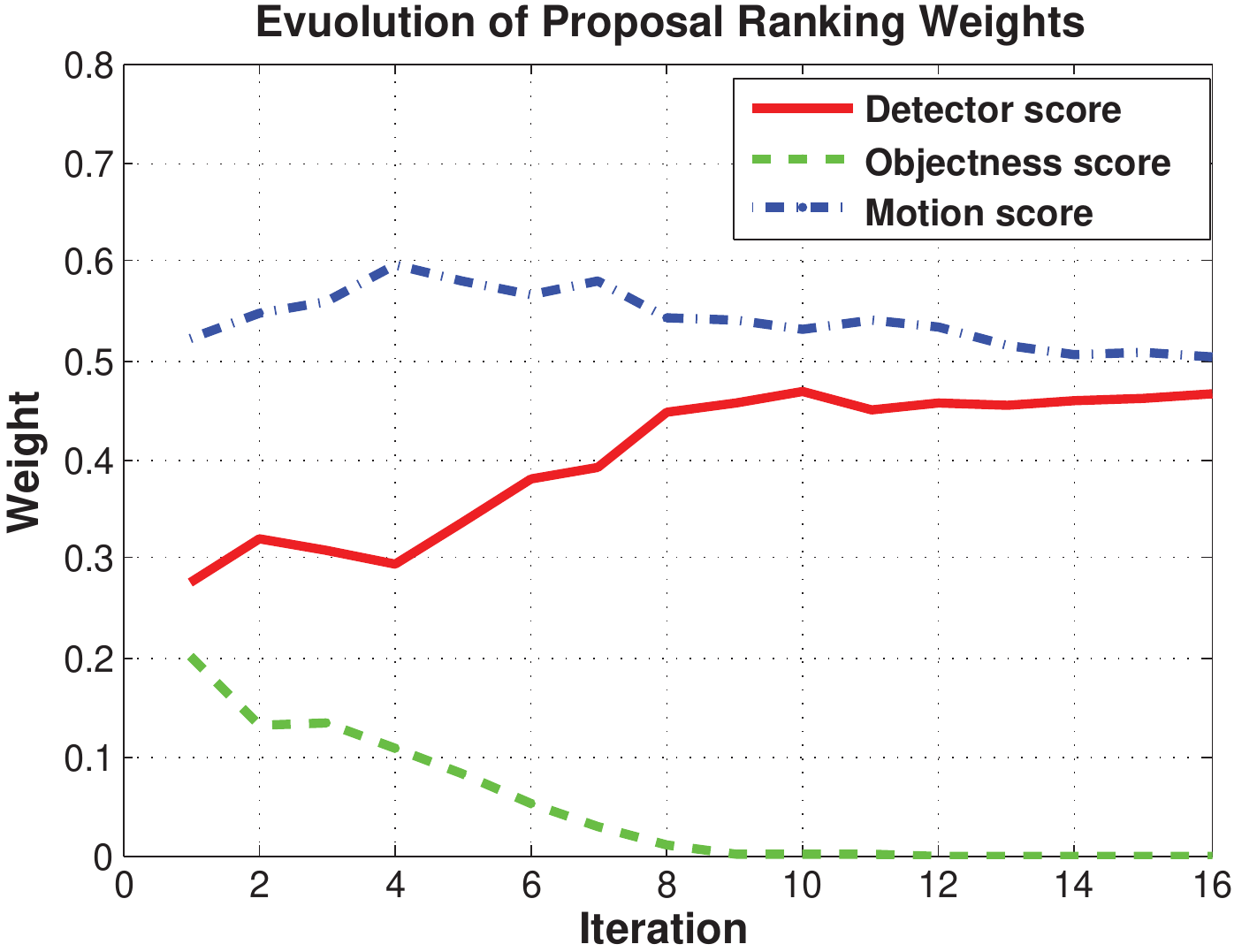}
\end{minipage}
}
\caption{\label{Fig.5} Validation of learning stability. (a) Monotonical decrease of sample error rates. (b) Evolution of proposal ranking weights.}
\end{figure}

\subsection{Model Effect}

In Fig.\ \ref{Fig.4}a and Fig.\ \ref{Fig.4}b, we respectively evaluate the effects of object enforcement and label propagation, showing that the PLM is more effective than the conventional LSVM model.

\textbf{Object enforcement:} Considering that the objective function in Eq. \ref{Eq.3} is non-convex, learning tends to get stuck into local minimum in the optimization procedure. By using the object enforcement procedure, Eq. \ref{Eq.5}, the performance of the learned detector significantly improved, Fig.\ \ref{Fig.5}a. The reason is that pedestrians are more precisely localised and most falsely detected object parts are depressed.  Given the 0.7 recall rate, the precision improved more than 10\% when using such a regularization term, which shows that the convex objective function does help the non-convex optimization to escape from poor local minimum.

\textbf{Label propagation:} Combined with the proposal ranking strategy, label propagation can incrementally annotate pedestrian samples without supervision. Fig.\ \ref{Fig.5}b clearly shows that the detection model is iteratively improved, showing the effectiveness of the graph-prorogation based incremental learning. After tens of iterations of learning, no additional positives are labeled and the performance is observed to be stable.

\textbf{Stability:}
Fig.\ \ref{Fig.5}a shows that the error rates of labeled training samples basically monotonically decreased, showing the stability of the proposed self-learning approach.
Fig.\ \ref{Fig.5}b shows the evolution of proposal ranking weights in the learning procedure of the PETS2009 dataset. The weight for the objectness score quickly decays to zero, which implies that the objectness score is not as discriminative as the detection and the motion scores. The weight for the detection score keeps increasing in learning, which indicates that the detector is progressively improved. The weight for motion cue decreases to a value that is similar to the detection cue, which implies that the motion feature is also discriminative.

Tab.\ \ref{Table1} shows the largest $\gamma$ values for the four datasets. $\gamma$ of the Towncenter dataset is the largest, while $\gamma$ of the CUHK dataset is the smallest. Larger $\gamma$ implies that the object proposals have fewer noises. The Towncenter dataset is a video with little illumination variance and few moving distracters, and therefore use a larger $\gamma$. The CUHK and 24Hours datasets have many moving distracters, so they need a smaller $\gamma$.

\begin{table}\footnotesize
\label{Table1}
\caption{Label propagation parameters on different datasets.}
\begin{center}
\begin{tabular}{|l|c|c|c|c|c|}
\hline
Dataset & PETS &Towncenter&PNN &CUHK&24Hours\\
\hline\hline
$\gamma$  & 0.50 & 0.70 & 0.60 &0.30 &0.30\\
\hline
\end{tabular}
\end{center}
\end{table}


\subsection{Performance}

\begin{figure}
\centering
\subfigure[]{
\begin{minipage}[b]{0.22\textwidth}
\includegraphics[width=1\textwidth,height=0.85\textwidth]{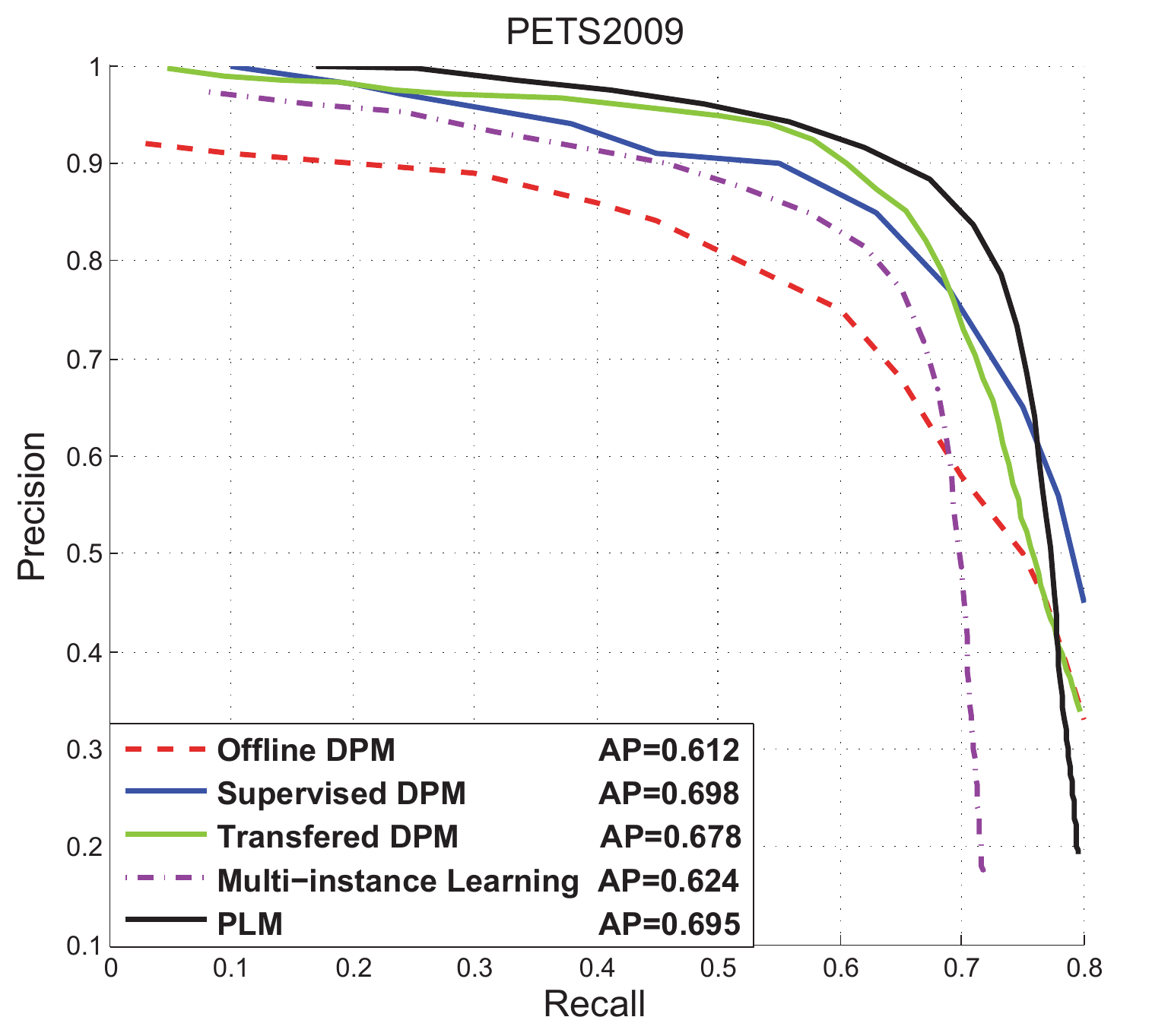}
\end{minipage}
}
\subfigure[]{
\begin{minipage}[b]{0.22\textwidth}
\includegraphics[width=1\textwidth,height=0.85\textwidth]{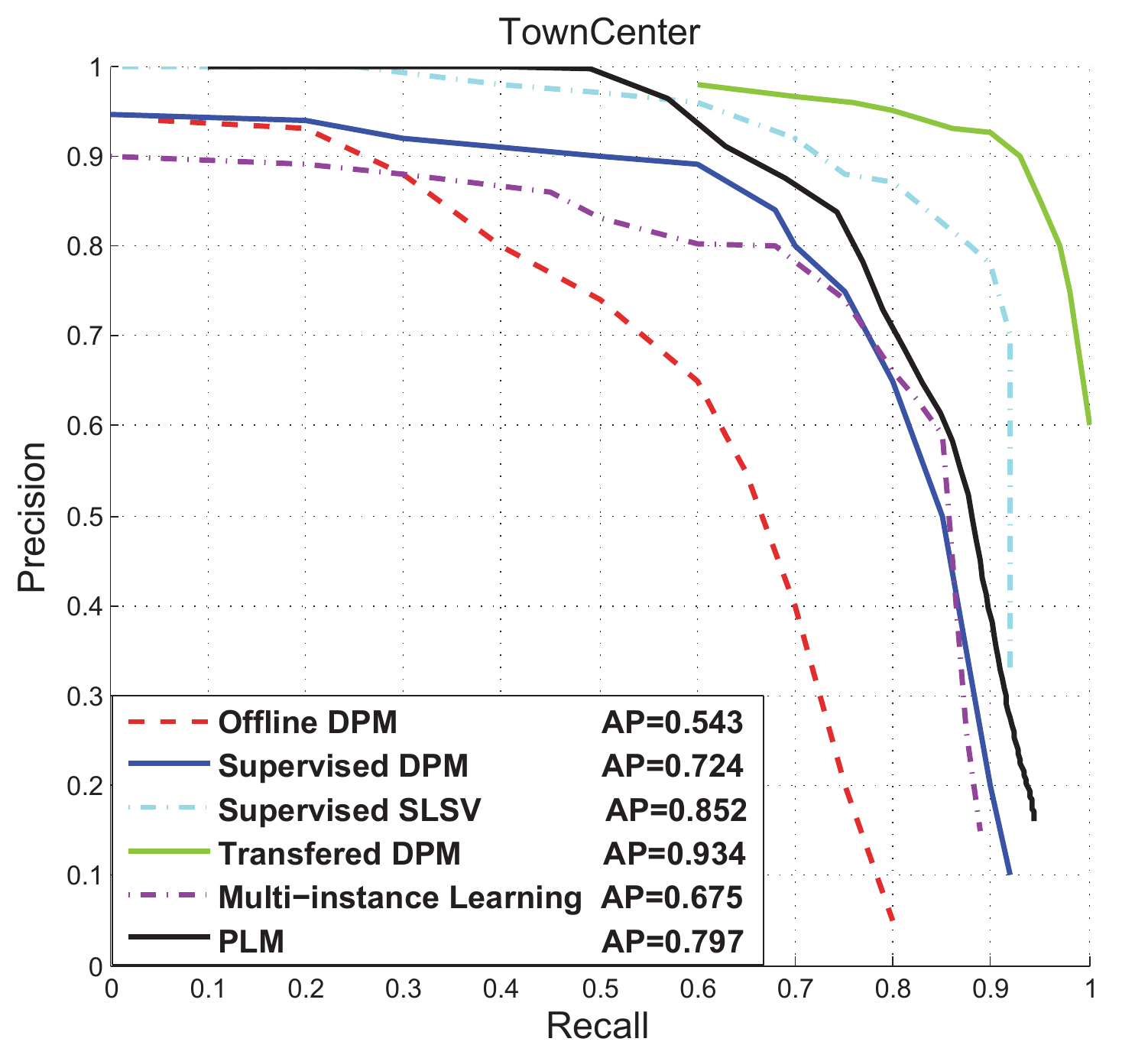}
\end{minipage}
}
\subfigure[]{
\begin{minipage}[b]{0.22\textwidth}
\includegraphics[width=1\textwidth,height=0.85\textwidth]{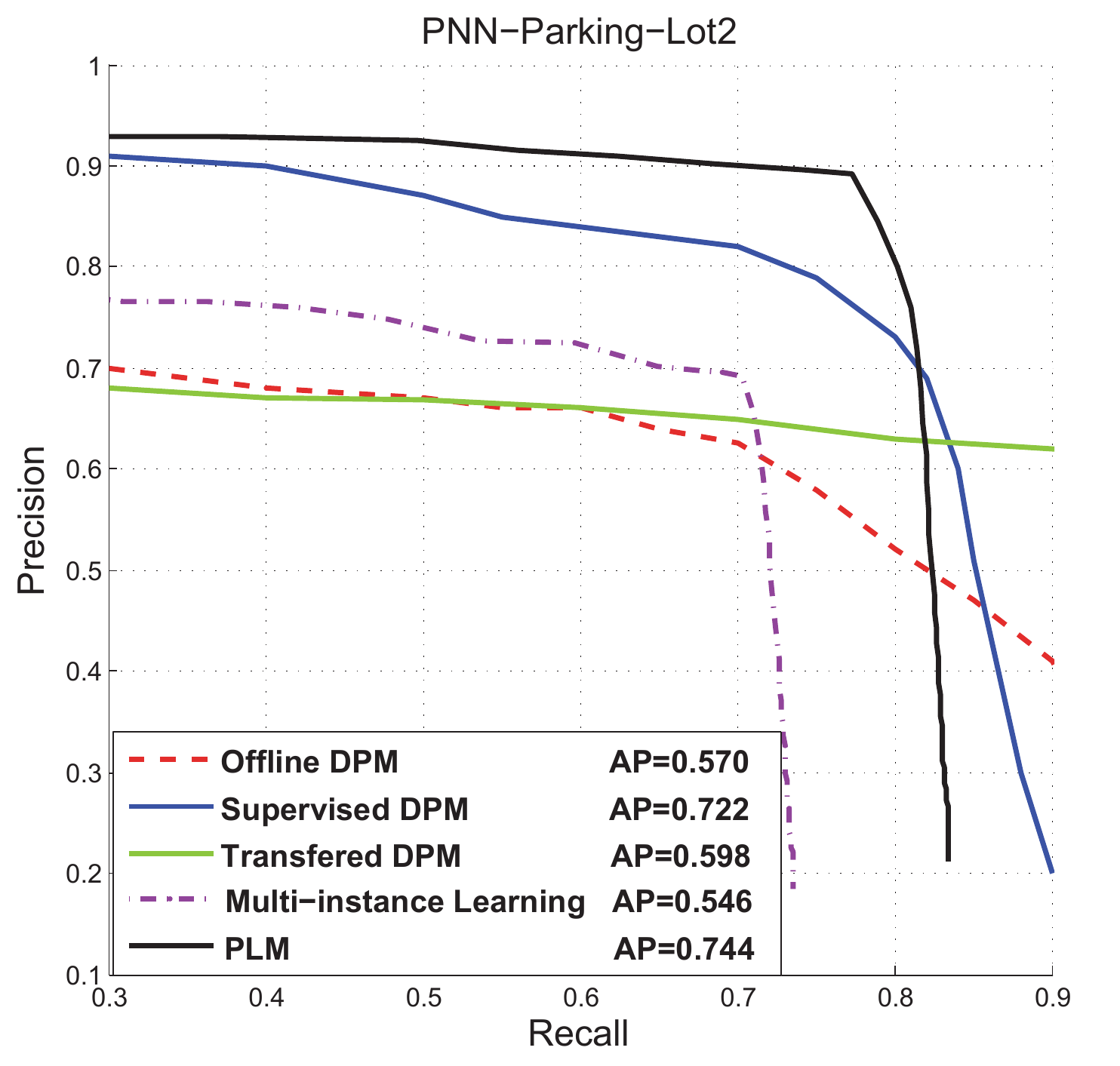}
\end{minipage}
}
\subfigure[]{
\begin{minipage}[b]{0.22\textwidth}
\includegraphics[width=1\textwidth,height=0.85\textwidth]{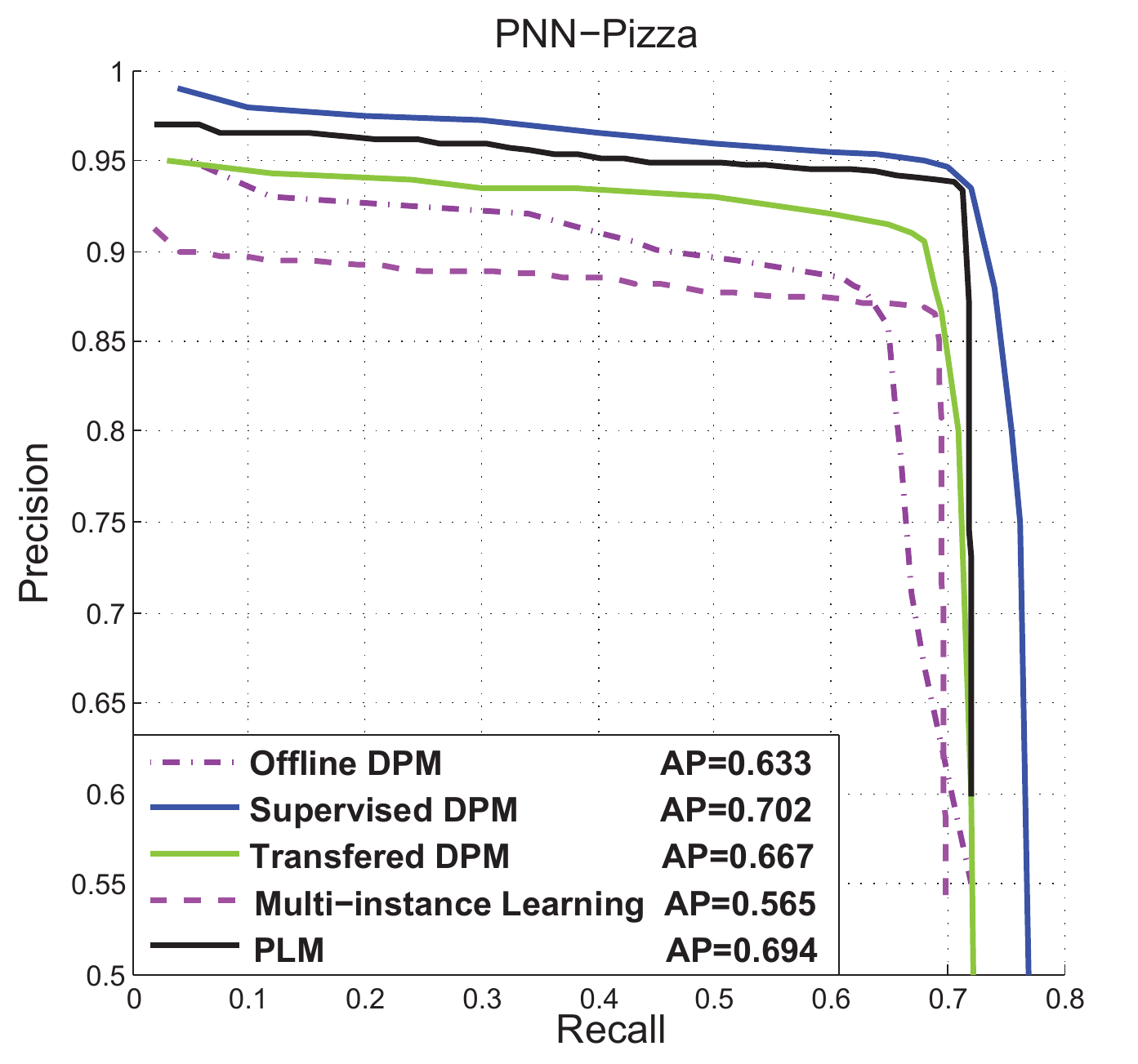}
\end{minipage}
}
\subfigure[]{
\begin{minipage}[b]{0.22\textwidth}
\includegraphics[width=1\textwidth,height=0.85\textwidth]{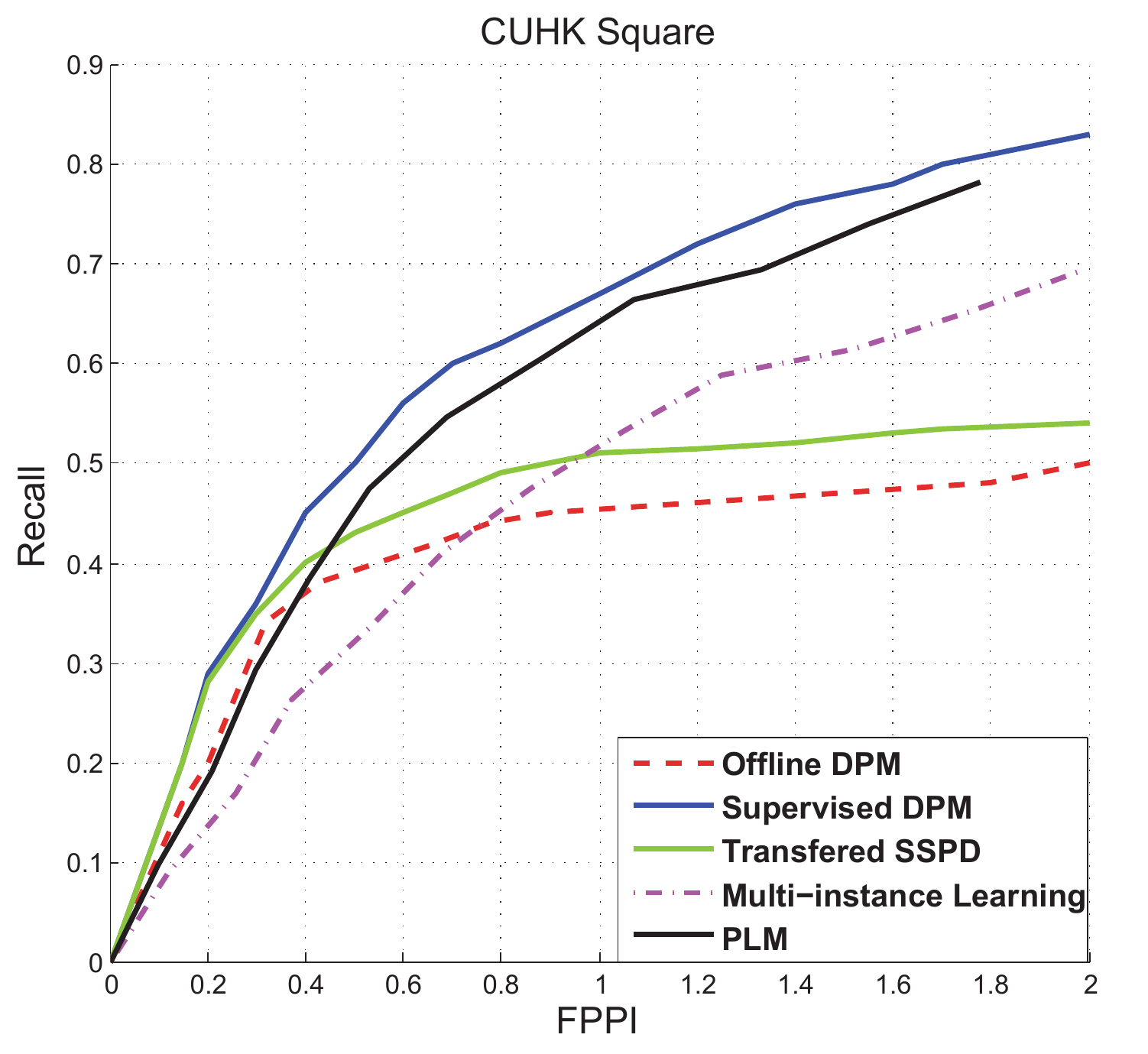}
\end{minipage}
}
\subfigure[]{
\begin{minipage}[b]{0.22\textwidth}
\includegraphics[width=1\textwidth,height=0.85\textwidth]{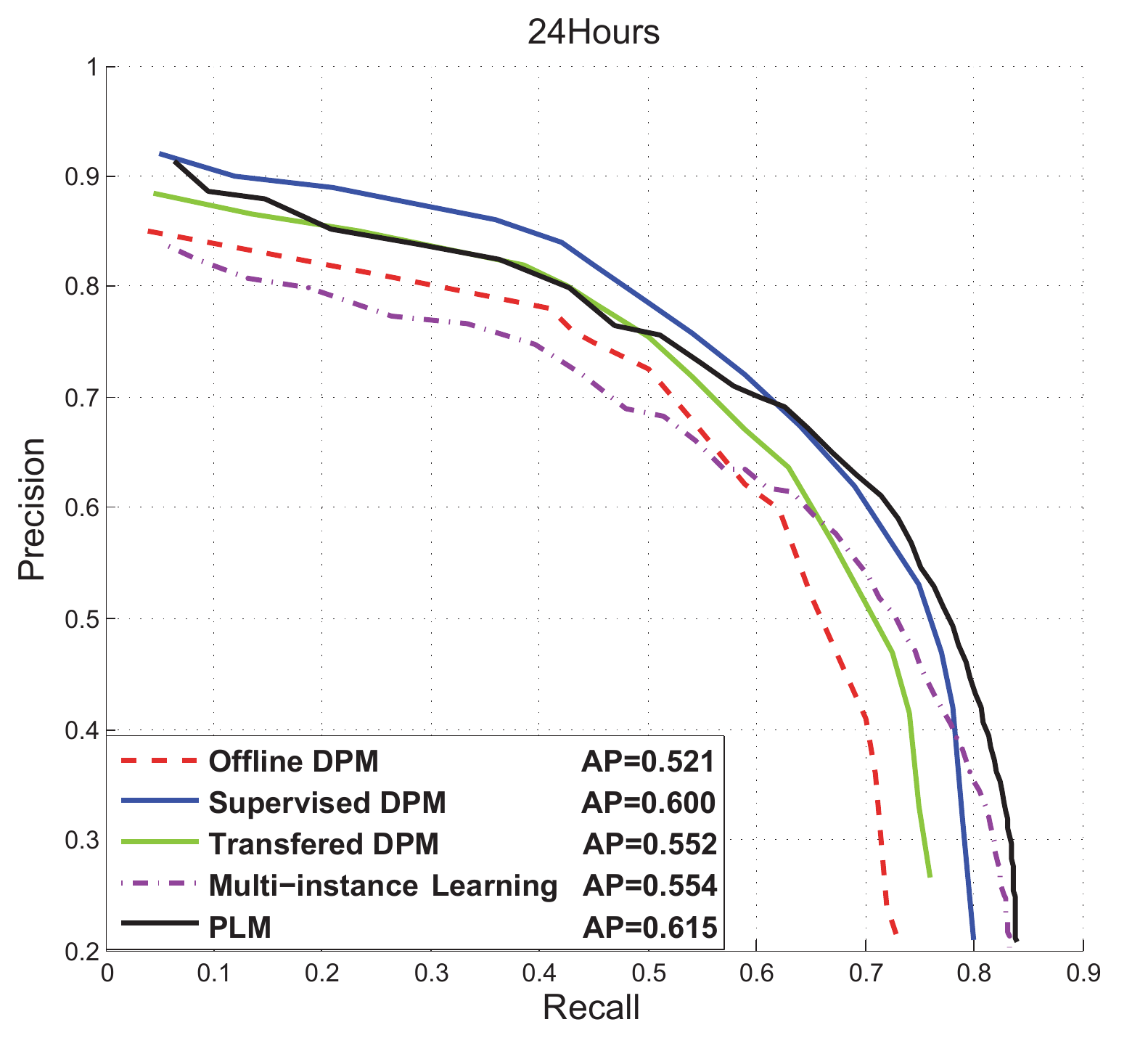}
\end{minipage}
}\caption{\label{Fig.6} Performance of our approach and comparisons with weakly supervised, supervised, and transfer learning approaches. On five datasets the Precision-Recall metric is adopted to evaluate the approach and compare it with other approaches. On the CUHK dataset the FPPI-Recall metric is adopted, consistent with the state-of-the-art scene-specific detection approach~\cite{Wang14}.
}
\end{figure}

\begin{figure*}
\centering
Pets2009 (crowd)\\
\includegraphics[width=0.98\textwidth,height=0.12\textwidth]{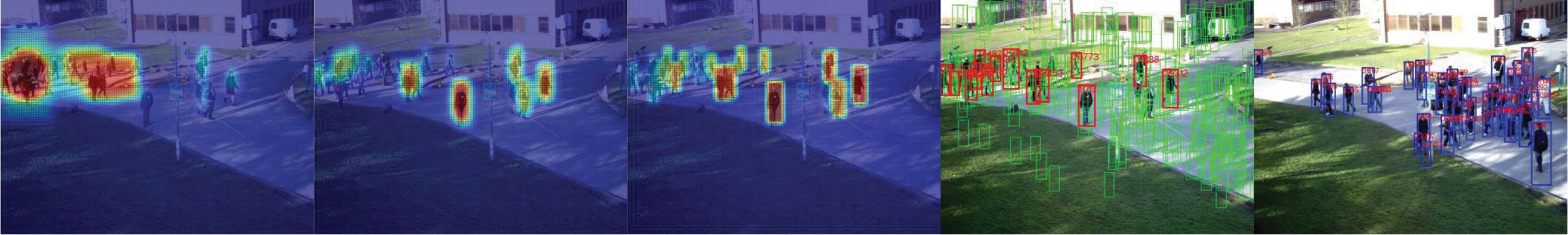}\\
Towncenter (moving distracters)\\
\includegraphics[width=0.98\textwidth,height=0.12\textwidth]{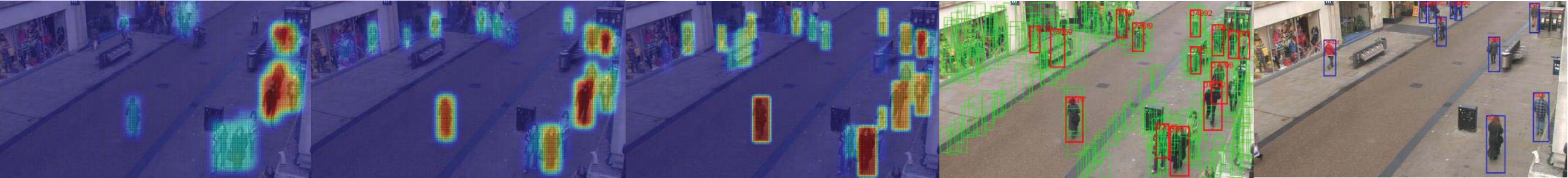}\\
PNN-Parking-Lots2\\
\includegraphics[width=0.98\textwidth,height=0.12\textwidth]{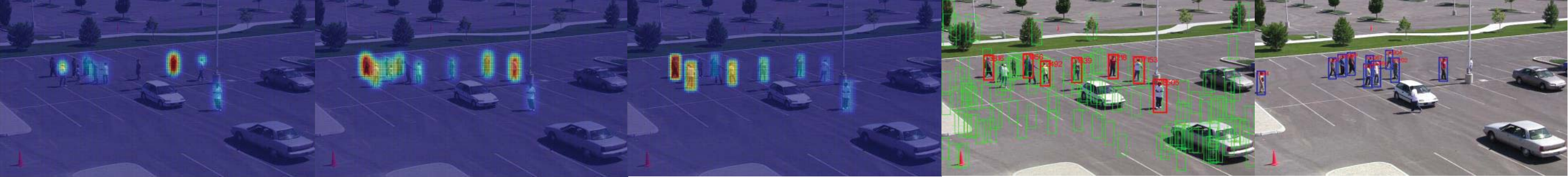}\\
PNN-Pizza(crowd)\\
\includegraphics[width=0.98\textwidth,height=0.12\textwidth]{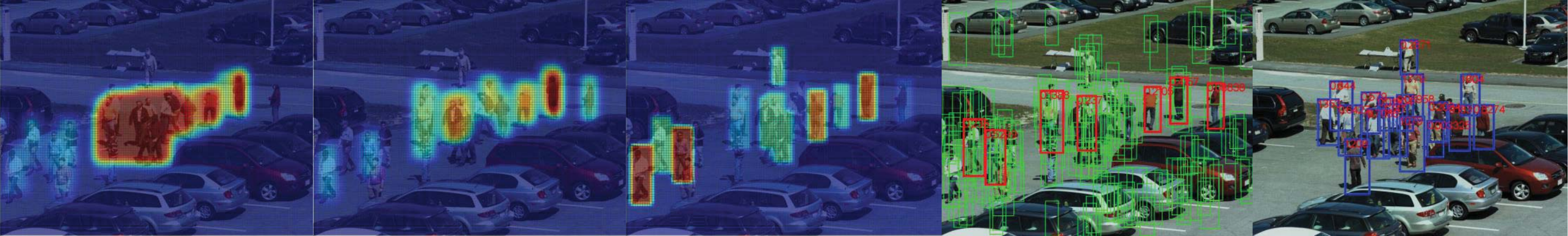}\\
CUHK Square (low resolution video with moving distracters)\\
\includegraphics[width=0.98\textwidth,height=0.12\textwidth]{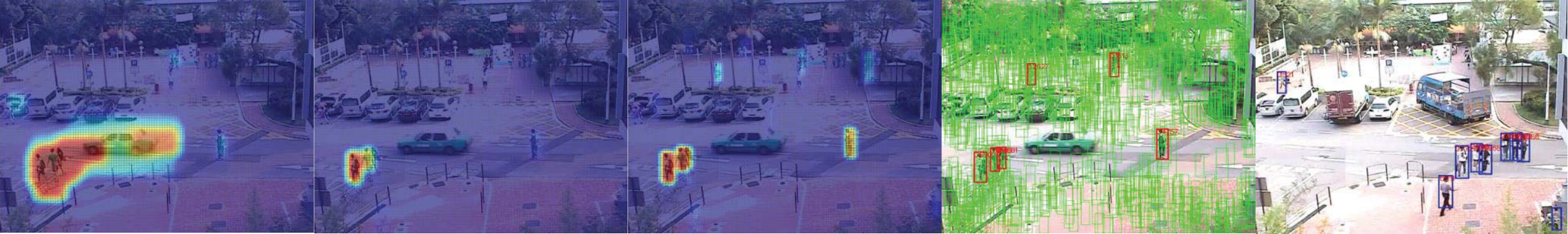}\\
24Hours(long video with moving distracters))\\
\includegraphics[width=0.98\textwidth,height=0.12\textwidth]{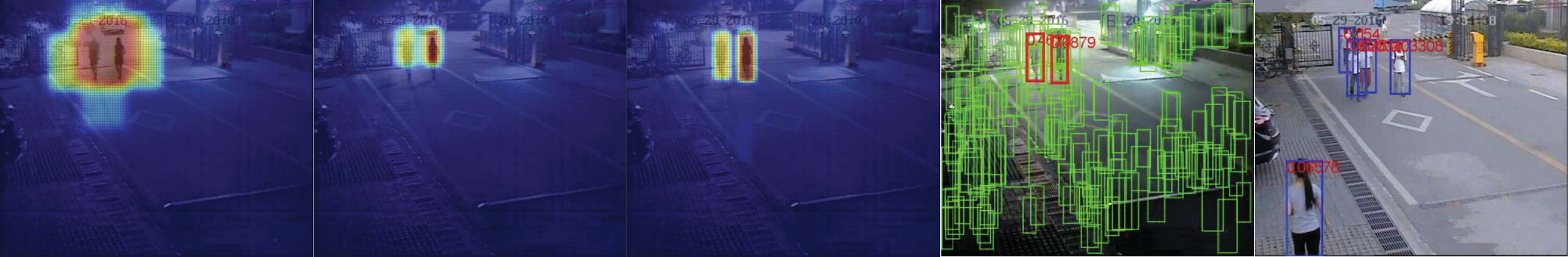}
\caption{\label{Fig.7} Illustration of learning and detection. First three columns: score maps in the first, firth and tenth learning iterations, respectively. Fourth column: annotated positive samples (red boxes). Last column: detection examples in the test sets. (Best viewed in color)}

\end{figure*}

The PR and FR curves in Fig.\ \ref{Fig.7} show that our approach significantly outperforms the off-line learned DPM detector on all datasets. It also significantly outperforms the Weakly-MIL approach. On the PETS2009 and PNN-Parking-Lot2 datasets, our approach outperforms all of the compared approaches. On the CUHK dataset our approach significantly outperforms the scene-specific approach with transfer learning ~\cite{Wang14}, which reports the state-of-the-art performance on this dataset. It is even comparable to the supervised learning approach (Supervised-DPM). On the Towncenter dataset, our approach outperforms the MIL approach as well. However, it shows lower performance than the fully supervised approach SLSV ~\cite{Hatori15} and the transfer learning approach ~\cite{Shu13}. The reason could be that the pedestrians in that video scene are sparse, thus our approach could not label sufficient positive samples. It should be stressed once again that our proposed approach does not use any annotated training sample.

On the 24Hours dataset, the AP (average precision) of our approach is highest among all compared approaches, Fig.\ \ref{Fig.7}e. It is about 6\% higher than the transfer learning method, validating our previous analysis: transfer learning suffers from the concept gap problem, e.g., adapt a model trained on day-time captured images to a video sequence of 24-hours illumination changes. By contrast, the proposed self-learning approach just applies the learned detectors from the same scenes, naturally avoiding the concept gap problem. More surprisingly, using additional motion cues, the proposed approach outperforms the fully supervised approaches in this dataset.

\begin{figure*}[!t]
\centering
\includegraphics[width=0.98\textwidth]{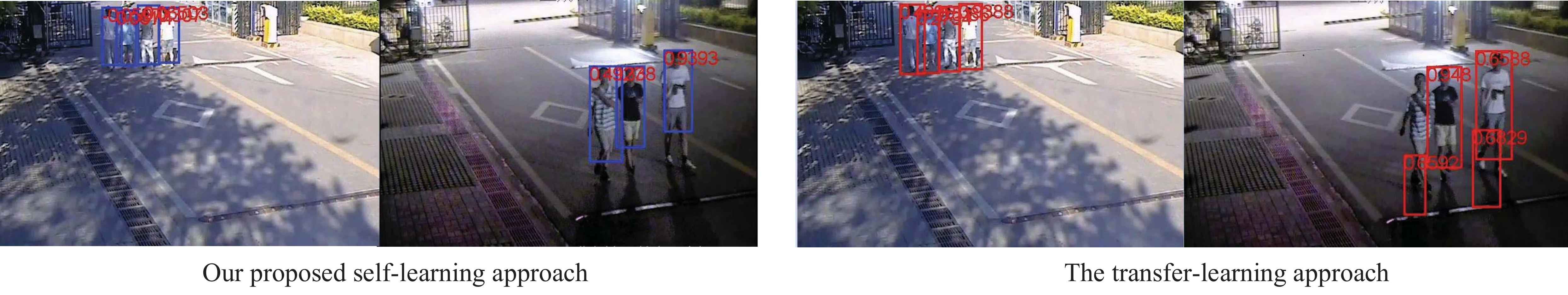}
\caption{\label{Fig.8} Detection results on 24Hours dataset. The self-learning detection correctly detects all pedestrians from the daytime (left) and night (right), but transfer learning has missed and false detections.}
\end{figure*}

In Fig.\ \ref{Fig.7}, we use key frames in each row to illustrate the incremental learning procedure.
It can be seen that the positive samples are incrementally labeled and noise samples are reduced. On the crowded PES2009 dataset and the PNN-Pizza dataset of significant occlusions our approach accurately labels samples, demonstrating that the learned detector has incorporated scene-specific discriminative information. On the Towncenter and CUHK datasets, although there exist moving distractors, e.g., bicycles and vehicles, the proposed approach correctly localize the pedestrians, demonstrating its robustness in noisy environments. In the 24Hours dataset, some video frames have dense pedestrians (daytime) but others have sparse pedestrians (at night). Learning from the early morning to the middle of the night, our approach could progressively improve its performance, without model drift. In the last column of Fig.\ \ref{Fig.7}, the detection results show that the learned scene-specific detectors are discriminative, showing robustness to occlusions, low resolution, and appearance variations. In Fig.\ \ref{Fig.8}, it can be seen that the self-learning approach is adaptive to view variance and 24-hours illumination changes, but transfer leaning suffers from those.

\section{Conclusions}
Supervised learning of detectors for all scenes requires significant human effort on sample annotation. Commonly used transfer learning and semi-supervised learning do not eliminate human supervision, as they require partial object-level annotations. We show that by leveraging extremely weakly annotated video data, it is possible to automatically learn customized pedestrian detectors for specific scenes. A new progressive latent model is proposed by incorporating discriminative and incremental functions. A self-learning approach is implemented by optimizing the model over spatio-temporal proposals. Experiments demonstrated that the self-learned detectors are comparable to supervised ones, taking a step towards self-learning cameras ~\cite{Gaidon14}.

\section*{Acknowledgement}
The partial support of this work by ONR, NGA, ARO, NSF, NSFC under Grant 61271433 and 61671427, and Beijing Municipal Science \& Technology Commission under Grant Z161100001616005 is gratefully acknowledged.

{\small
\bibliographystyle{ieee}
\bibliography{egbib}
}

\end{document}